\documentclass{llncs}
\usepackage{makeidx}  
\usepackage{graphics,graphicx,color,colortbl}
\usepackage{graphicx,subfigure,import}
\usepackage{float}
\usepackage{amssymb}

\usepackage{tabularx}

\usepackage[section]{placeins}

\begin{document}
\frontmatter          
\pagestyle{headings}  
%
%
\title{Automatic Localization of Deep Stimulation Electrodes Using Trajectory-Based segmentation Approach}
\titlerunning{Segmentation Electrodes DBS}  
%
\author{Roger Gomez Nieto \and Andres Marino Alvarez Mesa \and 
Julian David Echeverry Correa \and Alvaro Angel Orozco Gutierrez.}
\authorrunning{Roger Gómez Nieto et al.} 
%
%
%
\institute{Faculty of Engineering, Universidad Tecnol\'ogica de Pereira,\\ Pereira 660003, Colombia\\
\email{\{rogergomez,andres.alvarez1,jde,aaog\}@utp.edu.co}
}
\maketitle              
\vspace{-0.5cm}
\begin{abstract}
Parkinson's disease (PD) is a degenerative condition of the nervous system, which manifests itself primarily as muscle stiffness, hypokinesia, bradykinesia, and tremor. In patients suffering from advanced stages of PD, Deep Brain Stimulation neurosurgery (DBS) is the best alternative to medical treatment, especially when they become tolerant to the drugs. This surgery produces a neuronal activity, a result from electrical stimulation, whose quantification is known as Volume of Tissue Activated (VTA). To locate correctly the VTA in the cerebral volume space, one should be aware exactly the location of the tip of the DBS electrodes, as well as their spatial projection.\\ 

In this paper, we automatically locate DBS electrodes using a threshold-based medical imaging segmentation methodology, determining the optimal value of this threshold adaptively. The proposed methodology allows the localization of DBS electrodes in Computed Tomography (CT) images, with high noise tolerance, using automatic threshold detection methods.   

\end{abstract}
\vspace{-0.99cm}
\section{Introduction}
\vspace{-0.4cm}
Parkinson's disease (PD) is a degenerative condition of the nervous system, which manifests itself primarily as muscle stiffness, hypokinesia, bradykinesia, and tremor. In patients suffering from advanced stages of PD, Deep Brain Stimulation (DBS) neurosurgery is the best alternative to medical treatment, especially when they become tolerant to the drugs. 
DBS as a surgical treatment for PD symptoms was approved in 1998 by the United States Food and Drug Administration
. This surgery involves the implantation of a medical device which consists of three components: Implanted Pulse Generator (IPG), DBS electrodes, containing one to four platinum iridium contacts, and electric conduction cables. These three elements are implanted surgically into the human body \cite{Choi2011}.\\

Research on DBS is currently being carried out in various centers around the world, and there is an extensive published scientific literature on the subject \cite{Hemm2010}, \cite{Wardell2012}. DBS surgery produces a neuronal activity as a result of electrical stimulation generated by the electrodes. The quantification of neuronal activity is known as Volume of Tissue Activated (VTA) and can be interpreted as the amount of brain tissue that presents excitation or electrical response to stimulation generated by DBS electrodes. This VTA is a function of the DBS electrodes location within the brain. Therefore, the information regarding the position of these electrodes is used for the tuning of the stimulation.


Studies in patients with PD treated with DBS surgery show that the precision in the location of the electrodes within the desired anatomical region is directly correlated with motor enhancement during DBS \cite{Frankemolle2010}, \cite{Welter2014}. Although other factors influence the clinical results of DBS \cite{Lumsden2013}, the wrong location of the contacts is considered the most common cause of poor clinical response \cite{Horn2015}
. Consequently, knowledge about the location of the contacts becomes critical information in clinical routines to evaluate the effects of DBS and may help to define the best contact of the electrode that should be stimulated to treat chronic DBS, especially in the majority of current electrode designs, which have multiple stimulation contacts.


The location of electrodes used in DBS is still a challenging procedure, due to the presence of metal artifacts in the electrode and system cables for DBS \cite{Hebb2010}, \cite{Silva2015}. Dykstra et al. \cite{Dykstra2012} worked on approach with DBS electrodes, performing contact coordinates extraction by visual inspection of post-surgical CT. However, this method is limited to a reduced number of brain areas, due to the time required for visual extraction. Lalys et al. \cite{Lalys2013} worked on identifying optimal sites for DBS, constructing an atlas from post-surgical images. For the segmentation of electrodes used in the DBS, they applied a fixed intensity threshold, whose value is not specified, to extract the lower voxel of the patient electrodes. Lalys obtained an error of 1.31 mm while finding the barycenter of the voxel extracted over a 10 mm extension, which then allowed them to get the location of the contacts using the geometric constraints of Medtronic's 3389 electrode. However, they did not use actual electrode images for this location, but rather modeled the electrode axis within the atlas, using regression methods.

Motevakel and Medvedev \cite{Motevakel2014} used a technique called ``connect method'', which takes advantage of the linear shape of the electrodes and light beams reflecting the metal elements, resulting in a maximum error of 2.5 mm in the location of the electrodes used in DBS. However, their technique required that the two electrodes were present, which does not make them useful in the clinical cases where only one DBS electrode is introduced. To solve this limitation, some authors perform an artifact analysis for the electrode, based on semiautomatic algorithms \cite{Lalys2014}. Silva et al. \cite{Silva2015} used a cloud-based tool \cite{Tafula2014} to estimate automatically the position of the DBS electrodes compared with anatomical structures, using an estimation of the rectilinear trajectory of the DBS electrode, based on the upper and lower tip. For the location of each cranial electrode, they used the distances in mm from each contact with the bottom tip, specified by the manufacturer. Horn et al. \cite{Horn2015} determine the location of the electrode, through a semi-automatic process. They develop a Toolbox in MATLAB, which it has a precision range in the path location of two mm. Finally, they find the electrode using a straight-line model.


The medical image segmentation is a field of study that has been active for the last few decades, and results exist where the threshold varies proportionally to the average of the intensity of the CT \cite{Horn2015}. The methodology here proposed to determine intrinsic parameters of each CT potentially allow the segmentation to be adjusted to each patient and to have resulted not so generic but adjusted to the noise, distortion and other variables that vary between each CT.\\

In this paper, we propose a methodology for automatic segmentation of the DBS electrodes, which adjust to the intensity and rotation parameters of each CT scan. This method includes the CT image register with a clearly delimited anatomical atlas, which allows the creation of a Region of Interest (ROI) fully adjustable to the anatomy of each patient, followed by an intensity value that acts as a threshold for the segmentation, which is performed in this ROI. The performance of this methodology is measured by the comparison with a Ground Truth, using a Euclidean distance.

\section{Materials and Methods}
\label{Materials_Methods}
\begin{figure}[]
	\centering
	\includegraphics[trim={0cm 0cm 0cm 0cm},clip,width=0.99\linewidth]{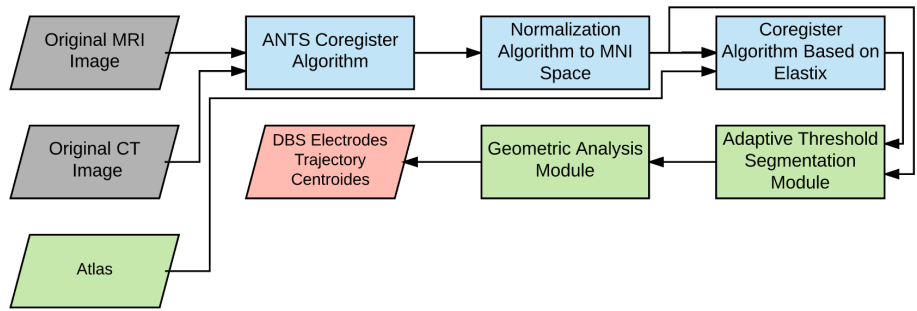}
	\caption{General diagram of the proposed methodology. Rectangle frames represents processes in which information is transformed. Parallelogram represent inputs or outputs of information.}
	\label{fig:Diagrama_General_Metodologia}
\end{figure}

This section presents the materials and methods used for the development of the proposed methodology. The Figure \ref{fig:Diagrama_General_Metodologia} shows the general diagram, where the gray boxes represent the medical images as inputs, blue boxes represent state of the art techniques proposed by others authors, and green boxes represent the new contributions made it. Finally, the red box represents the centroids defining the trajectory of DBS Electrodes, as main output from the method.  

\vspace{-0.3cm}
\subsection{Atlas Originated}

 Due to significant variations in orientation in different CT images, which will depend on from the CT scanner in which they are performed, a single ROI cannot be generated for all patients. Therefore, this research proposes an iterative and autonomous method, where an ATLAS is created for a patient, defining a region of optimal and precise interest for that patient, determined by well-defined anatomical coordinates. This ROI should cover most of the DBS electrodes trajectory but should minimize the presence of bone structures within it.\\ 
 
 initially, to determine a geometric approximation of this ROI, the dimensions of the electrode trajectory in MNI space are taken into account, as shown in Table \ref{Table:Dimensiones_Trayectoria_Electrodo_MNI}. This lists the dimensions of the electrode trajectory in the patients analyzed, so it provides significant geometrical constraints to generate the ROI defined by the coregister with Atlas.   
 
 \begin{table}
 	\centering
 	\caption{Dimensions of the electrode trajectory in MNI space. These distances are calculated on the axial axis, taking into account the slice where the electrode tip is located and the slice where the DBS electrodes touch the patient's skull.}
 	\label{Table:Dimensiones_Trayectoria_Electrodo_MNI}
 	\begin{tabular}{c|c|c|c|c}
 	 	\hline
 	 	Patient & Electrode  & Electrode & Absolute      & Volume size\\
 	 	ID      & start (mm) & end (mm)  & distance (mm) & axial (mm)\\
 	 	\hline
 	 	P1			&	-36		&	48.5	&	84.5	&	114\\
 	 	P2			&	-11		&	51.5	&	62.5	&	198\\
 	 	P3			&	-22.5	& 	43		&	65.5	& 	168\\	
 	 	P4			&	-50		&	40		&	90		&	150\\
 	 	P5			&	-41.5	&	20.5	&	62		&	162\\
 	 	P6			&	2.5		&	71		&	73.5	&	156\\
 	 	P7			&	-12.5	&	66		&	78.5	&	221\\
 	 	\hline
 	 	Mean Vale 	& -24.429	&	48.643	&	73.786	& 167	\\
 	 	\hline	
 	\end{tabular}
 \end{table}
 
\subsection{Coregister Atlas and CT MNI}

For this coregister, which represents the third system of Figure \ref{fig:Diagrama_General_Metodologia}, the software Elastix \cite{MetzKleinSchaapEtAl2011} is used. The MRI image registration process is performed by a linear process with a rigid register, and then a register based on an affine transformation.

Let $\Psi=\left\lbrace  x_{r} \in \mathbb{R}:\mathbf{r}=(i,j,k)\right\rbrace $ be an structured set of intensity voxels of image CT normalized to MNI space, where $x_{r}$ is the Hounsfield intensity measured at location $\mathbf{r}\in \mathbf{R} \subset \mathbb{N}^{3}$.


\begin{equation}
Y=\left\lbrace  x_{r} \in R^{s\times a}:r \in {1, \cdots, a }\right\rbrace\ .
\end{equation}

Where $r$ is the number of slices will be analyzed. Let $\mathbf{\hat{Y}}$ be an ROI, obtained by multiply;

\begin{equation}
\mathbf{\hat{Y}}=\mathbf{Y}\circ \mathbf{Z}\ .
\end{equation}

Where the function $\mathbf{Z}$ is an ROI defined as $\mathbf{Z} \in [0,1]^{M \times N \times O}$.Finally, we obtained a binary image $\mathbf{B}$, defined as:

\begin{equation}
\end{equation}

   \begin{equation}
\mathbf{B_{r}} = \left\{
\begin{array}{ll}
1 & \mathrm{if\ } \mathbf{Y(x)} \circ \mathbf{Z(x)} \geq T_{r} \\
0,    & otherwise.
\end{array}
\right.
\end{equation}

Where $T_{r}$, is the threshold calculated for each slice in the axial axis.

\subsection{Adaptive threshold}

The electrode trajectory was identified using threshold segmentation. For this, we used the approximation by Rodriguez et al. \cite{Rodriguez2006} and Otsu et al. \cite{Otsu1975}, where they describe the Local Adaptive Thresholding method. Because there exist a significant level of variation in the intensity of the gray-scale level within the range of $ 0-4000$ HU, the global threshold does not work well here. Local threshold works best for this case where the image is divided into slices, so we find the single threshold for each slice. The threshold value will depend on regional statistics such as variance and mean of the image. We get the threshold $T$ by adding to the intensity average $\mu$ of the slice in the CT image, the amount $K \times \sigma$ of the Eq. \ref{Eq_Formula_Threshold_Horn}:   

\begin{equation}
T=[K \cdot  \sigma]+\mu\ .
\label{Eq_Formula_Threshold_Horn}
\end{equation}

Where $\sigma$ and $\mu$ are, respectively, the standard deviation and mean of intensities values of pixels, within the local window. $K$ is a variable that can be determined to obtain the best segmentation. However, approaching the threshold in this way, with the proposed value for $K=0.9$ by \cite{Horn2015}, it can be shown that all or most of the voxel intensity values of DBS electrodes can be calculated below the threshold, which will not allow a correct segmentation. To find the optimal value of K parameter that provides the best segmentation, we next performed an exploratory analysis, where K value is varied by of $± 0.01$, in the range $[0.6 - 9.5]$. After was evaluating the Euclidean distance that defines the segmentation for each of these values of K. The upper limit for the range is determined in $9.5$; since from 9.5, the error initiates an upward trend, and this is a fact that is repeated in patients of the database.   

\vspace{-0.3cm}
\subsection{Validation}
\vspace{-0.1cm}
Because it is not desired to compare the volume of the electrode, but only its centroid, we use Euclidean distance, which establishes the separation distance between the centroid of Ground Truth, and the centroid calculated with the proposed methodology. To compare the results with the results obtained in \cite{Horn2015}, we must decrease the resolution of the planned trajectory. So, the trajectory with a resolution of $0.5$ mm is again sampled so that in the end it has a resolution of $0.63$ mm (LEAD). For this step, we used the concept of linear interpolation. 

\subsection{Dataset} 
The dataset was collected at the Instituto de Epilepsia y Parkinson del Eje and Instituto Neurol\'ogico de Colombia. It is composed of eight (8) computed tomography scans, applied to an equal number of patients, all diagnosed by a specialist as patients with Parkinson's Disease. These imaging studies by Tomography have an average resolution of 1.45 mm. In order to get the Ground Truth, the centroids of the electrodes are manually labeled in the slices containing electrodes. In this study, were considered more than 1600 by hand labeled centroids, belonging to slices that contain electrodes.\\ 
\section{Results and Discussion}
\label{Results}

\begin{figure}[t] 
	\centering 
	\includegraphics[trim={4cm 1cm 4.7cm 0.5cm},clip,width=0.99\linewidth]{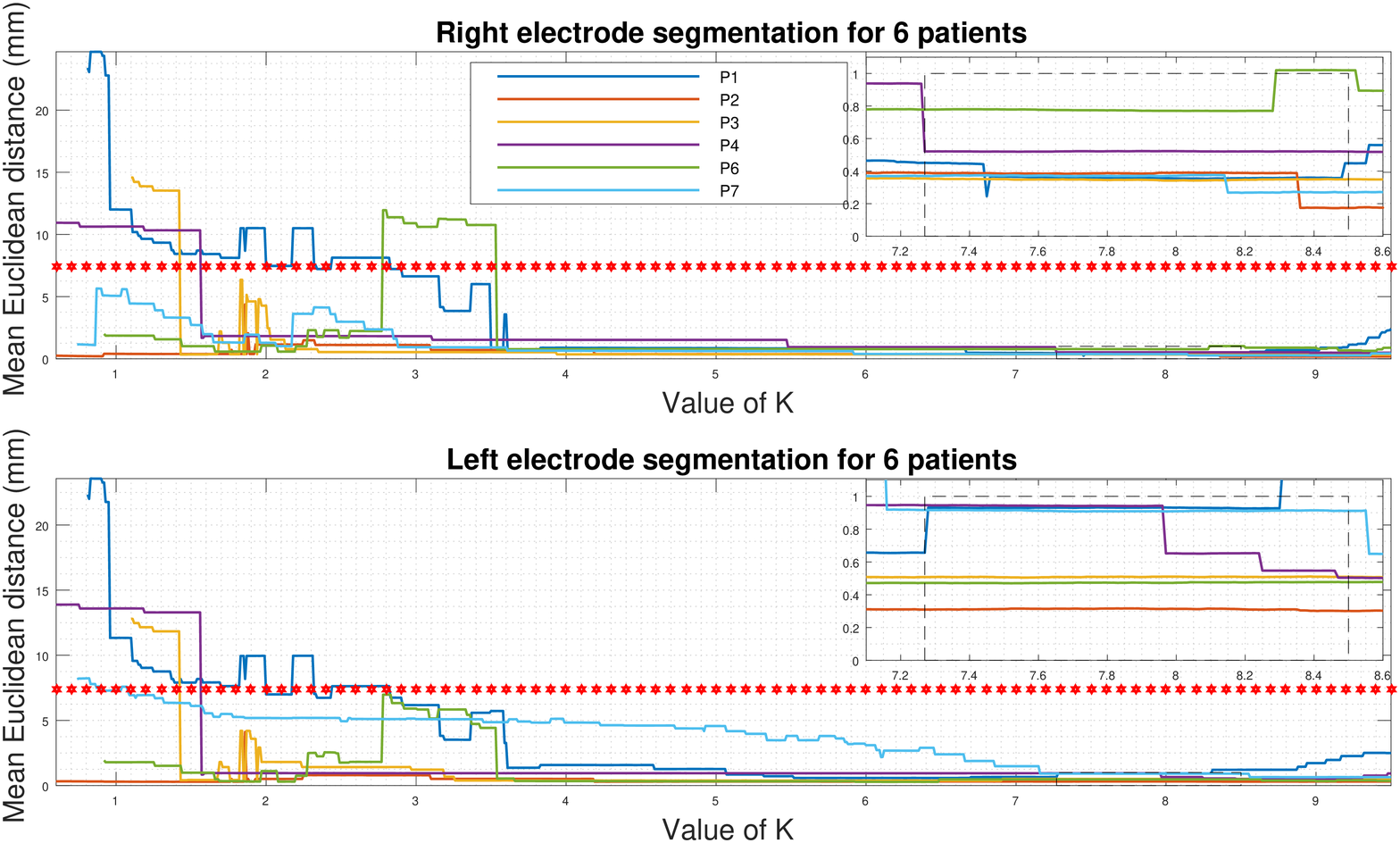} 
	\caption{The Euclidean distance that defines the DBS segmentation results of electrodes in 6 patients (P1,P2,...), the increase of K in each iteration is 0.01 mm. The stars (\textit{red}) line shows the Euclidean distance obtained by the Toolbox LEAD; this distance is only valid for K = 0.9; however, for comparison purposes, an extension is made over the range of K, marked with red stars. The frame with the black dotted line represents the K-scale values found with the proposed methodology, in this interval, the asset value of Euclidean distance that defines the segmentation is less than one mm in all the training patients. A zoom of this range is performed.} 
	\label{fig:Distancia_Euclideana_Segmentacion} 
\end{figure}

In Figure \ref{fig:Distancia_Euclideana_Segmentacion} is shown the mean Euclidean distance (ED) that defines the segmentation accuracy for six patients, using the proposed method for K values from the range [$0.6-9.8$]. The ED presented by LEAD, on average, for the six patients, is shown with red stars. This distance for LEAD is calculated only at K = 0.9. However, an extension to this line is made, using red stars, to show that the ED that generates the method, is consistently inferior to the ED of LEAD.\\ 
In more than 50 $\%$ patients, the methodology has a lower Euclidean distance, even with the same K value as the one used in LEAD, then showing the improvement in the precision provided by the proposed method. The unique conditions as noise and intensity for each patient made that a single K value cannot be defined as the only appropriate value, which will be able to use for all patients, we found that a qualitative analysis of the range should be performed. In this qualitative study, a range of optimal K values is determined, for which the Euclidean distance is minimized for the database patients, thus finding the best possible general segmentation.\\ 

From Figure \ref{fig:Distancia_Euclideana_Segmentacion} we can deduce that all K values higher than four, have a lower Euclidean distance than LEAD. However, to make more precise the location of the K range, it is placed as a condition that all K values within this range should allow a segmentation whose Euclidean distance with Ground Truth is less than $1$ mm for all patients. Finally, we obtain that the range of values for K selected for the methodology is \textbf{[$7.27-8.5$]}.\\   

\subsection{Proposed Atlas for coregister}

Due to significant variations in orientation in different CT images, which will depend on from the CT scanner in which they are performed, a single ROI cannot be generated for all patients. Therefore, this research proposes an iterative and autonomous method, where an ATLAS is created for a patient, generating a region of optimal and precise interest for that patient, determined by well-defined anatomical coordinates. This ROI should cover most of the DBS electrodes trajectory but should minimize the presence of bone structures within it.\\ 

Then the Atlas for the ROI coregister is defined anatomically, precisely and easily reproducible, as follows: 

\begin{enumerate} 
	\item In the sagittal plane, there are no variations; the only limit is the range [$-40, +40$] mm. 
	\item In the caudal part, the axial axis has values of [-49, -39.5] mm, while the coronal axis is limited between [-30, +3] mm. 
	\item In the immediately upper part, the axial axis is limited in the range [-39, +1.5] mm and the coronal axis is limited between [-30, +22] mm. 
	\item In the most cranial part of the Atlas, the axial axis is limited in the range [$+2, +31$] mm and the coronal axis is limited between [-30, +32] mm. 
\end{enumerate}

\begin{figure}[]	
	\centering     
	\subfigure[Atlas for adaptive ROI generation.]{\label{fig_Atlas}\includegraphics[trim={0cm 0cm 0cm 0cm},clip,width=0.46\linewidth]{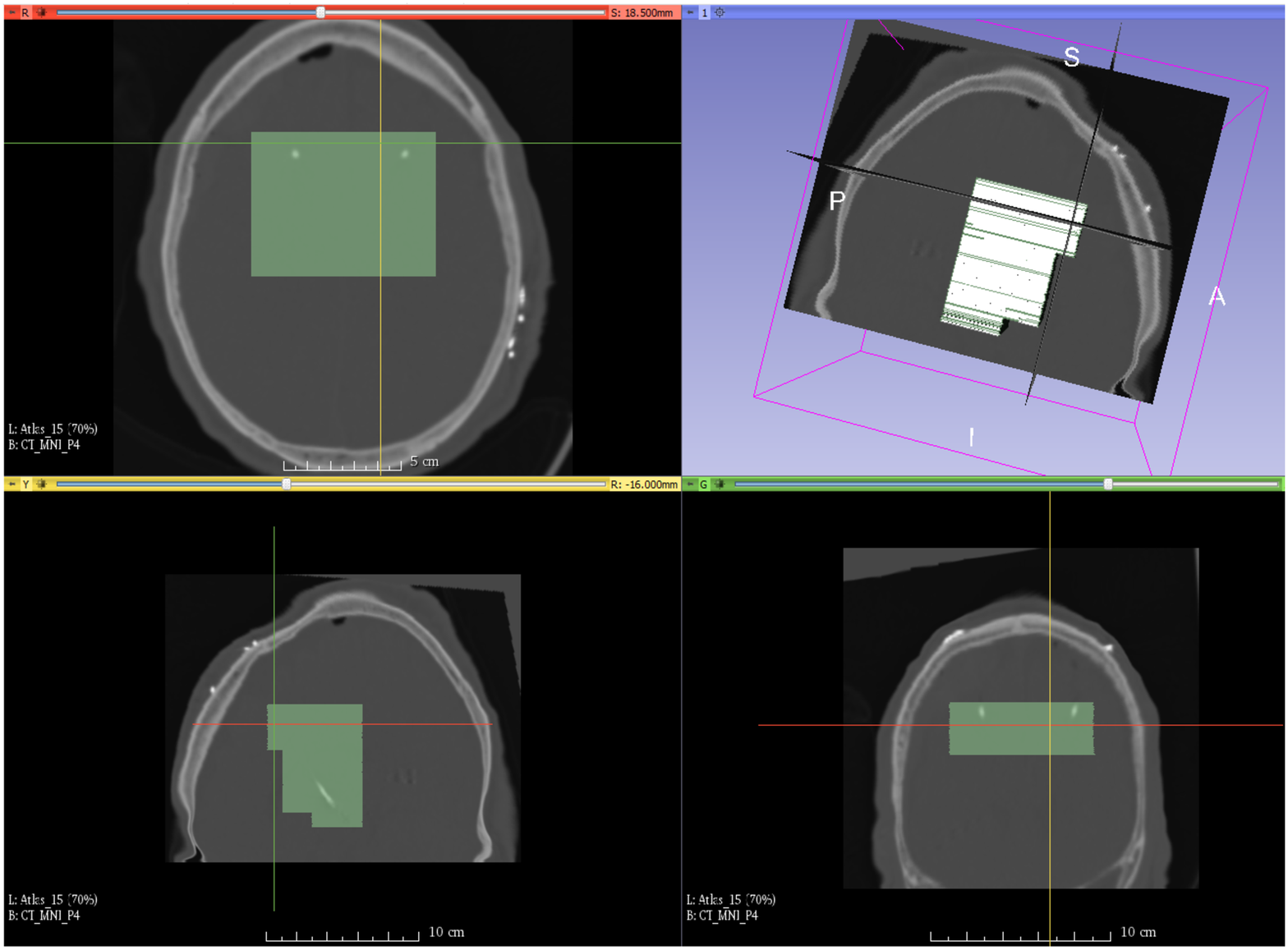}}
	\subfigure[ROI generated by co-registration between the proposed atlas with the CT MNI of P3.]{\label{fig_Resultados_P3_Coregister}	\includegraphics[trim={0cm 0cm 0cm 0cm},clip,width=0.49\linewidth]{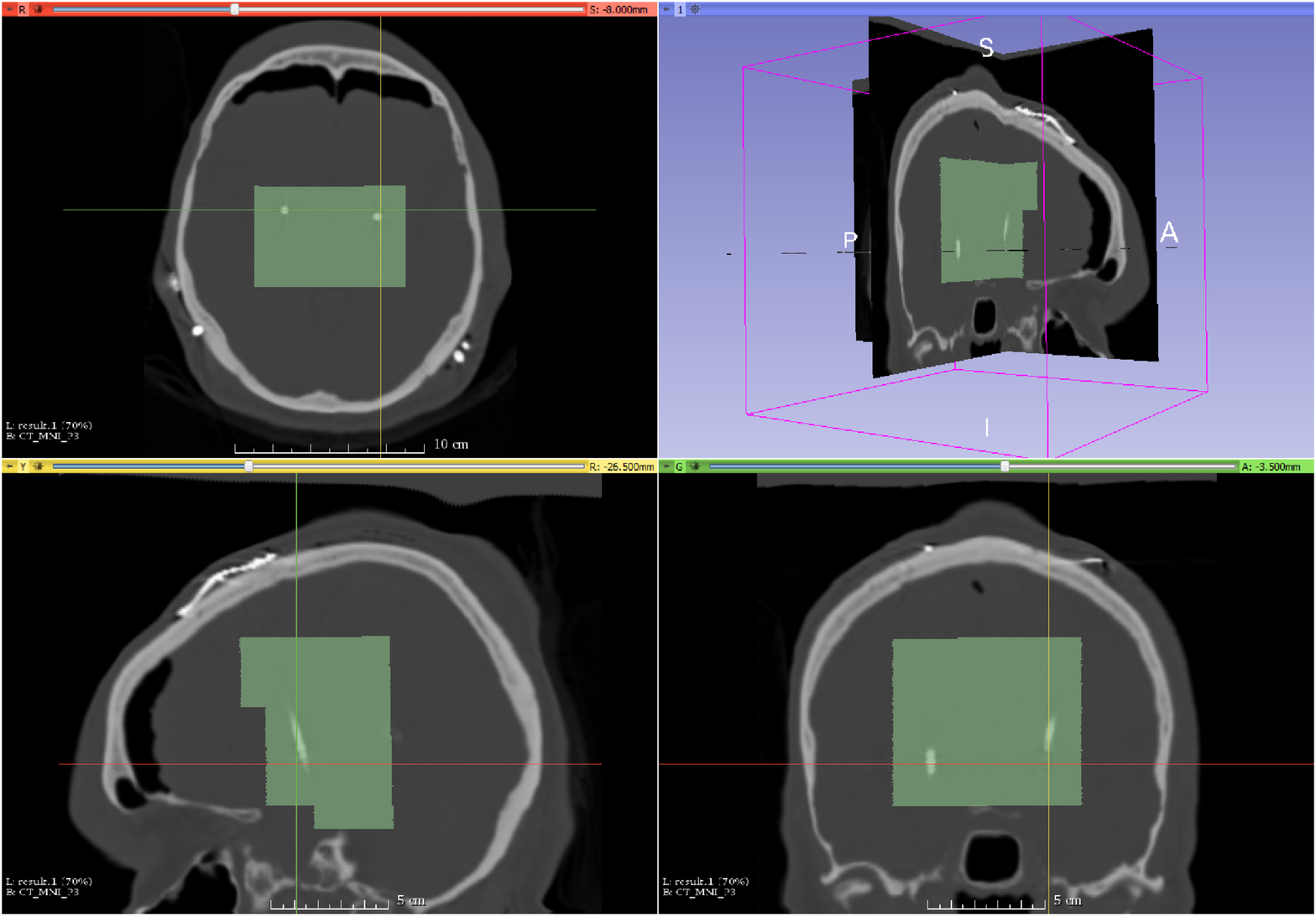}}
	\caption{Atlas for an adaptive ROI generation (\textit{Left}) and ROI originated by the coregister of this atlas with the MN3 CT of P3 (\textit{Right}). It is observed in the right image that the new ROI generated minimizes the inclusion of bone or cranial structures, thus producing a greater level of precision in the segmentation by an adaptive threshold. These images were obtained from the Slicer software.} 
	\label{Fig_Co-registro_ATLAS_P3} 
\end{figure} 

This ROI is 80 mm deep in axial axis, has a width of 80 mm in sagittal axis, and in coronal axis, the length changes because the electrode trajectory travels mainly in this axis. It was set up from Patient 4 CT, this because it is the patient who presents the greatest depth of the electrode in the axial axis. So, it is the patient who could introduce more significant measures of error and presence of cranial structures. In Figure \ref{Fig_Co-registro_ATLAS_P3} the resulting mha format image (green) is shown after the coregister between ATLAS and MNI P3 CT. It is observed that the atlas is well delimited, minimizing the cranial area in analysis, and allowing to cover most of the electrode trajectory.\\ 

In table \ref{Tabla_Errores} is shown the Mean Squared Error (MSE) between the trajectory coordinates of both electrodes, delivered by the proposed method and the LEAD method, and the Ground Truth coordinates. Last, the average and standard deviation of this error are reported. It is observed that the proposed method consistently has a reduced error compared to the error generated by the LEAD method. It should be noted this may be due to the LEAD method is not an automatic method, so a manual correction must be made in the height on the axial axis by the specialist. In the error measure presented here, there is no kind of height correction was performed, which explains the presence of an MSE of up to 207 mm in some patients for the LEAD method.\\

\begin{table}
	\centering 
	\caption{Mean Squared Error (MSE) between Ground Truth and the coordinates of electrode trajectory generated by LEAD and proposed Method. Last, the average value and standard deviation for each method are reported for both electrodes.}
	\label{Tabla_Errores}
	\begin{tabular}{c | c| c | c| c}
		\hline
		Patient ID & \multicolumn{2}{|c|}{Right Electrode} & \multicolumn{2}{c}{Left Electrode}\\
		& MSE OWN (mm) & MSE LEAD (mm) & MSE OWN (mm) & MSE LEAD\\
		\hline
		P1 & 0.049 & 81.68  & 0.35  & 54.38  \\
		P2 & 0.05  & 0.442  & 0.038 & 2.049  \\
		P3 & 0.04  & 0.91   & 0.1   & 0.83   \\
		P4 & 0.054 & 207.49 & 0.174 & 229.91 \\
		P6 & 0.24  & 3.13   & 0.081 & 3.93   \\
		P7 & 0.051 & 4.91   & 0.338 & 4.84   \\
		P8 & 0.448 & 29.307	& 0.197 & 125.12 \\
		\hline
		Mean $\pm$ SD & 0.134 $\pm$ 0.155& 46.838 $\pm$ 76.66 & 0.182 $\pm$ 0.122 & 60.151 $\pm$ 87.737\\
		\hline 
	\end{tabular}
\end{table}

\begin{figure}[]
	\centering     
	\subfigure[Patient 5]{\label{fig_Resultados_P3_Segmentation3D}\includegraphics[trim={0cm 0cm 0cm 0cm},clip,width=0.66\linewidth]{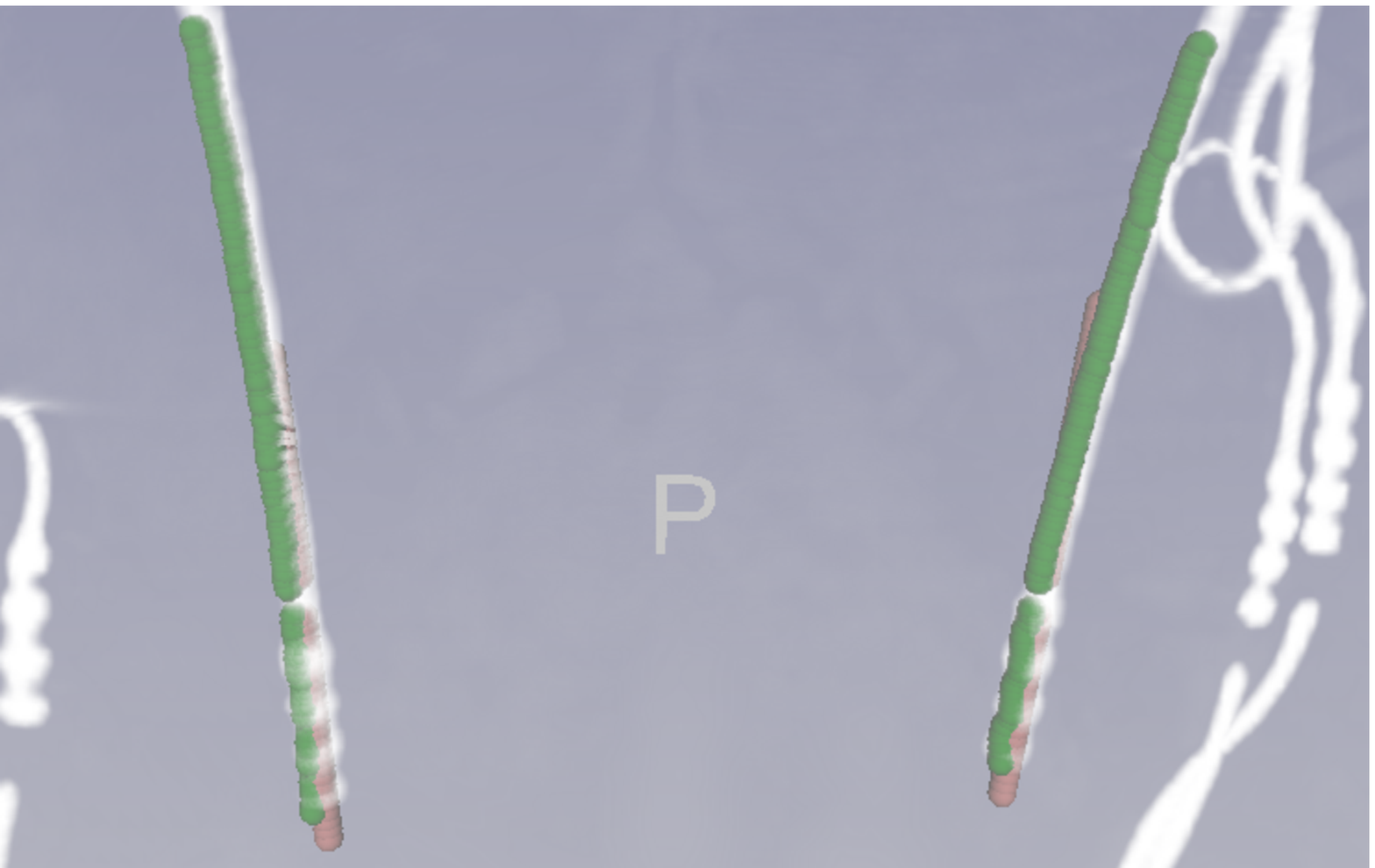}}
	\subfigure[Patient 8]{\label{fig_Resultados_P4_Segmentation3D}	\includegraphics[trim={0cm 0.0cm 0cm 0cm},clip,width=0.318\linewidth]{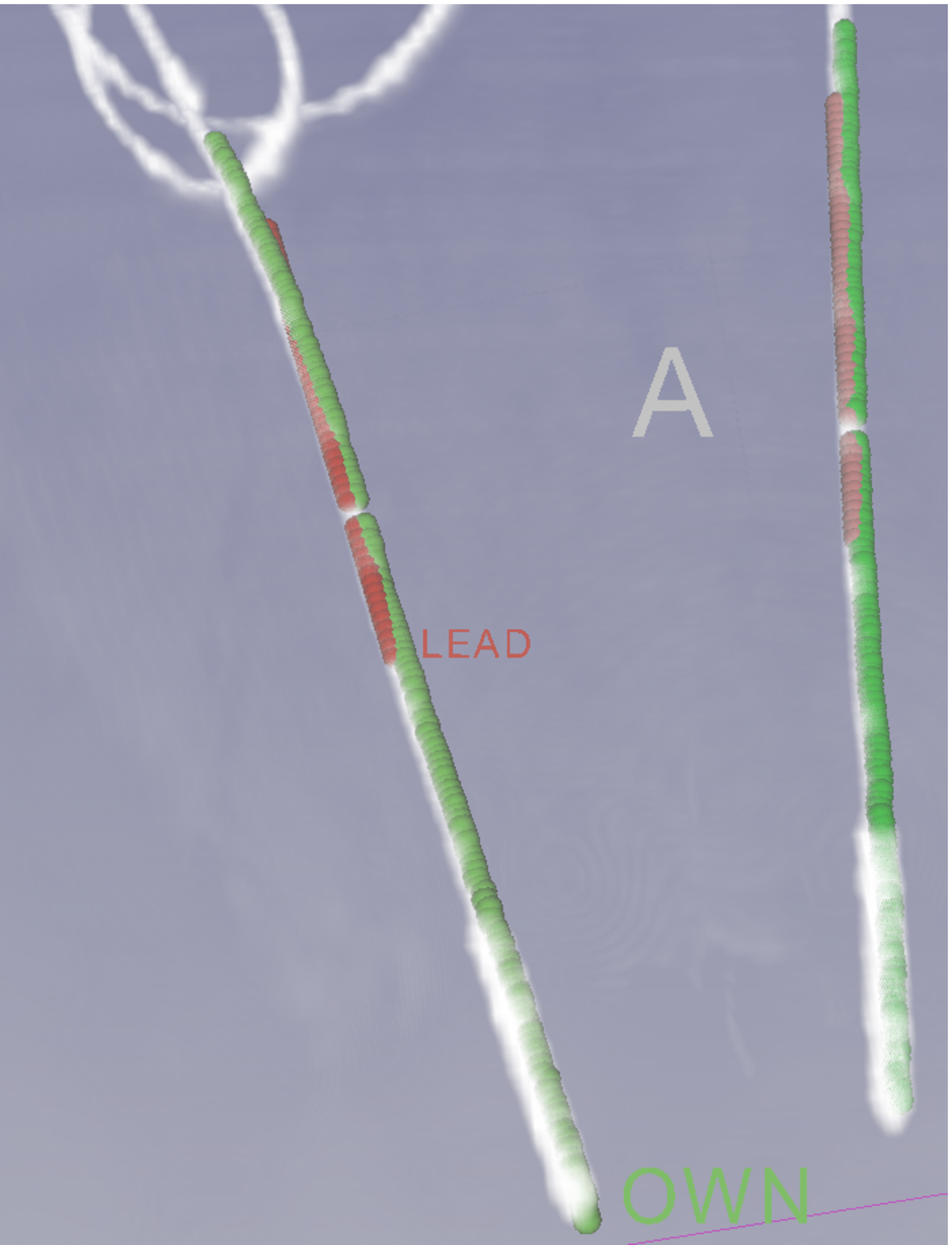}}
	\caption{3-D reconstruction of the original electrodes (\textit{white}) and the path found with the proposed methodology (green) and LEAD method (red), defined by semi-spherical volumes.}
	\label{fig:Reconstruccion3D}
\end{figure}

In Figure \ref{fig:Reconstruccion3D} is shown the 3D reconstruction of DBS electrode trajectory provided by the proposed and LEAD method placed over the image of the electrodes (blank). It is observed that the LEAD method (red) does not define well the electrode trajectory. The Figure \ref{fig_Resultados_P3_Segmentation3D} shows the best result obtained by the LEAD Method in the electrode segmentation and the figure \ref{fig_Resultados_P4_Segmentation3D} the worst case is shown. It is observed that although LEAD follows the trajectory correctly in the coronal and sagittal axes, it does not have concordance with the trajectory along the axial axis. This lack of accuracy of LEAD is because it is not an automatic method, so it requires a specialist adjustment in the axial axis depth. Instead, the proposed method segments the electrode trajectory automatically (green), without supervision or correction of any kind. The electrode trajectory obtained by applying this methodology matches with an accuracy of $0.2$ mm the right trajectory of the electrodes, thus preliminary validating our research.\\   

\begin{figure}[ht!]
	\centering
	\includegraphics[trim={0cm 0cm 0cm 0cm},clip,width=0.99\linewidth]{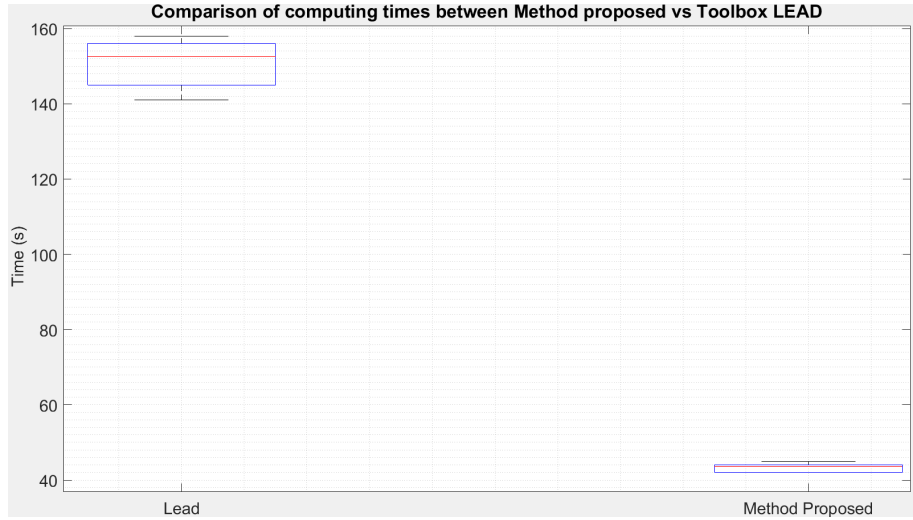}
	\caption{Comparison of the trajectory calculation times of the DBS electrodes between the Toolbox LEAD \cite{Horn2015} and the method proposed here.}
	\label{fig:Comparison_times}
\end{figure}

The Figure \ref {fig:Comparison_times} shows the time comparison of the segmentation process of the DBS electrodes, both in Toolbox LEAD as using the methodology proposed here. It is demonstrated that the proposed method takes around $28 \%$ of the time it takes LEAD to arrive at a result. That is, based on the computation times in Fig. \ref{fig:Comparison_times}, there is an improvement of $78$\% in the computational time taken by the method on the time it takes the Toolbox LEAD. The computational times were measured in a dual-core i5 2450-M 2.5 GHz processor, based on the Sandy Bridge architecture, with 12 GB of RAM, Intel HD Graphics 3000 integrated video card and 128 GB internal SSD.\\

The accuracy DBS electrodes segmentation is an important task, particularly since a spatial and geometric reference point for any tool that intends to generate information of physical or chemical processes involving these DBS electrodes. That is why this paper provides a new methodology that allows DBS electrodes segmentation to be performed with an accuracy of less than $0.2$ mm, in the eight patients in the database. Due to a limited number of patients for to validate the results, it is important to corroborate these results with larger databases, to verify the proper functioning of electrodes segmentation. Regarding state of the art, the present work makes a contribution by proposing a methodology that has proven to perform the segmentation of the DBS electrodes automatically and accurately. This advanced method suppresses the independence of specialist that have other tools since the electrodes trajectory are not required manual corrections in any of three anatomical axes.

\section{Conclusions}
\label{Conclusions}

In conclusion, we provide here a methodology of segmentation of medical images for automatic localization of electrodes used for the surgery of Deep Brain Stimulation, a procedure that is performed in the treatment of Parkinson’s disease. The methodology proposed has been shown to adapt to the patient’s conditions regarding noise, distortion and other parameters of the CT. Results indicate that the method can locate with an accuracy of less than 1.5 mm per slice, the effective path followed by the DBS electrodes, improving what is found in state of the art. 
The proposed method has positive clinical implications, such as that specialist should not provide feedback in the process of the DBS electrodes segmentation, since the method proposed here in the dataset used has proved to be automatic and accurate. Additionally, the computation time of DBS electrodes segmentation has dramatically decreased, reaching reductions of up to $78 \%$ in calculation time of the trajectory, compared with previous approaches.\\ 

Future work can be focused on: 
\begin{itemize} 
	\item \textbf{Future work 1.} Validate the proposed methodology with a larger number of patients, with the objective of patenting both the methodology and the software, for clinical use. 
	\item \textbf{Future work 2.} To implement the necessary changes in the source code that allow a parallel execution, using concurrent programming and by threads; all this to improve the execution time. 
	\item \textbf{Future work 3.} To implement the methodology proposed in free software, such as the ITK library, following a pipelining method, which allows executing all the computer libraries from a single software environment. 
	\item \textbf{Future work 4}: 
\end{itemize} 

\textbf{Acknowledgments}. This work was developed within the project 1110-657-40687 ``Estimaci\'on de los par\'ametros de neuro modulaci\'on con terapia de estimulaci\'on cerebral profunda en pacientes con enfermedad de Parkinson a partir del volumen de tejido activo planeado", with financial support of Colciencias (Colombia). Authors were also supported by the 706 agreement,  ``J\'ovenes Investigadores e Innovadores", with the proposal ``Desarrollo de una metodolog\'ia para localizaci\'on autom\'atica de electrodos utilizados en DBS (estimulaci\'on cerebral profunda) para el tratamiento de la enfermedad de Parkinson", funded by Colciencias (Colombia). Finally, the authors are thankful to the `Grupo de Investigaci\'on en Automática' ascribed to the engineering program at the Universidad Tecnol\'ogica de Pereira, and M.Sc. H.F. Garc\'ia for technical support.

%
\bibliographystyle{IEEEtran}
\bibliography{Biblio_IBPria2017}
\end{document}